# PCANet-II: When PCANet Meets the Second Order Pooling

Lei Tian, Xiaopeng Hong✉, Guoying Zhao,
Chunxiao Fan, Yue Ming, and Matti Pietikäinen

*Abstract*—PCANet, as one noticeable shallow network, employs the histogram representation for feature pooling. However, there are three main problems about this kind of pooling method. First, the histogram-based pooling method binarizes the feature maps and leads to inevitable discriminative information loss. Second, it is difficult to effectively combine other visual cues into a compact representation, because the simple concatenation of various visual cues leads to feature representation inefficiency. Third, the dimensionality of histogram-based output grows exponentially with the number of feature maps used. In order to overcome these problems, we propose a novel shallow network model, named as PCANet-II. Compared with the histogram-based output, the second order pooling not only provides more discriminative information by preserving both the magnitude and sign of convolutional responses, but also dramatically reduces the size of output features. Thus we combine the second order statistical pooling method with the shallow network, i.e., PCANet. Moreover, it is easy to combine other discriminative and robust cues by using the second order pooling. So we introduce the binary feature difference encoding scheme into our PCANet-II to further improve robustness. Experiments demonstrate the effectiveness and robustness of our proposed PCANet-II method.

## I. INTRODUCTION

FACE is one of the most interesting subjects in various computer vision tasks, and many face analysis tasks have achieved significant progress in recent years. However, there are still many unsolved problems in real applications, such as extreme intra-class variations and large number of subject classes in face recognition (FR) and affective computing (AC) tasks. In order to solve these problems, large number of learning-based methods [1][2], especially deep learning (DL) methods [3][4], have been proposed in recent years. The convolutional neural network (CNN) [5] is a typical deep learning model. Its architecture contains three main components: convolutional layers, activation functions and pooling layers. The FaceNet [3], which follows the idea of end-to-end learning, directly learns the mapping from raw image to Euclidean space. The work [4] integrates the center loss with original softmax loss functions to further enhance the deep features' discriminative power. However, these methods require a large number of samples to train the DL model. Comparatively, a number of "shallow" learning models are proposed, such as PCA-Network (PCANet) [6] and Stacked Image Descriptors (SID) [7]. Lei *et al.* [7] introduced several existing shallow descriptors into the deep model by stacking image descriptor layers and max-pooling layers alternatively. These shallow descriptors include Principal Component Analysis (PCA), Linear Discriminant Analysis (LDA), Discriminant Analysis with Tensor (DTA) [8] and Discriminant Face Descriptor (DFD) [1]. Another recent noticeable work, i.e., PCANet [6], achieved good performance for image classification. Compared with CNN, the PCANet does not need the backward propagation process to update the model's parameters, therefore, it performs well on small-scale dataset. The PCANet is considered as a simplified model of CNN [6]. We briefly review the overall process of PCANet for the integrity of our paper. The PCANet follows the basic architecture of CNN and consists of a few convolutional layers, non-linear processing layers and block histogram extraction layers. For convolutional layers, PCANet computes the local patch based filter kernels by the PCA. For non-linear processing layers, the whole feature map is binarized by a unit step hashing function. For the pooling layer, different feature maps are assigned to difference weights. Just like LBP [9], around each pixel, a set of binary values are summed with weights and a decimal-valued image is obtained. The block-wise histograms are computed in each local block from the decimal-valued image. At last, the block-wise histograms are concatenated into a long vector. We abbreviate the **block-wise histogram** as **histogram** hereinafter unless it is specifically indicated.

The PCANet method uses histogram technique as the pooling approach. However, there are three main problems for this kind of pooling methods. **Problem 1:** The binary hashing process sets pixel value of a feature map to be 1 when the original pixel values are larger than 0, otherwise, to be 0. This binarization process makes a lot of discriminative information loss during the pooling process. **Problem 2:** It is difficult for the histogram-based pooling method to combine other effective visual cues during the pooling process. **Problem 3:** The dimensionality $2^L$ of histogram grows exponentially with the number of feature maps $L$. Therefore, it limits the number of feature maps in the convolutional layer.

Lei Tian, Chunxiao Fan, Yue Ming are with Beijing University of Posts and Telecommunications, P.R.China (e-mail: tianlei189@bupt.edu.cn; fcxg100@163.com; myname35875235@126.com). Lei Tian, Xiaopeng Hong, Guoying Zhao and Matti Pietikäinen are with the Center for Machine Vision and Signal Analysis, University of Oulu, 90014, Finland (e-mail: Lei.Tian@oulu.fi; Xiaopeng.Hong@oulu.fi; guoying.zhao@oulu.fi; matti.pietikainen@oulu.fi).

In this paper, in order to solve the above problems, we use the second order statistics to pool the feature map set. Moreover, in order to improve the robustness of our method, we also introduce the binary feature difference (BFD) encoding scheme into our PCANet-II. We would like to not only obtain more discriminative information from the floating based response value of feature maps (w.r.t **Problem 1**), but also expect to preserve the robustness which is provided by the LBP-like pooling method of original PCANet. Benefiting from the expandability property of the covariance matrix, we can integrate these two excellent properties into our method simultaneously (w.r.t **Problem 2**). In other words, we encapsulate other pixel-wise responses into the floating feature maps by a covariance matrix. However, the pooling scheme of original PCANet produces a kind of discrete patterns' index instead of numerical features. It is meaningless to directly combine discrete index based response with the feature maps. Therefore, we employ the BFD scheme to translate the non-numerical pooling response to numerical response. The numerical difference responses can be encapsulated into the feature map set. Multiple kinds of information features enhance the discriminative power of our method. Moreover, the covariance matrix is a kind of statistics instead of the distribution variable. So the dimensionality of covariance matrix's output is relatively lower (w.r.t **Problem 3**).

There are several works about second order pooling for CNN model [10][11][12][13]. However, to our best knowledge, there is no second order pooling scheme for PCANet model. The methods based on CNN model need backward propagation process and lots of samples to update the parameters of network. In contrast, our PCANet-II learns model's parameters by one forward propagation, so it works well on small-scale dataset. Moreover, the computational cost of our PCANet-II is also dramatically reduced.

## II. THE PROPOSED APPROACH

In this paper, we propose a novel "shallow" network model with the second order pooling and BFD, namely PCANet-II. Fig. 1 illustrates the framework of our PCANet-II model. We first compute the filter kernels and convolutional responses of all stages. Next, for the convolutional responses of each stage, we compute the BFD images from a set of convolutional feature maps, and cumulate feature maps and BFD images together as a new feature map set. Then, the second order statistics is computed from the new feature map set of all stages. At last, we vectorize the covariance matrix and stack the output of all stages as the final output vector. The convolutional layer of our model is similar to the original PCANet. However the following steps of our model are totally different from the PCANet model. In the following sections, we will describe the computational details of each step.

### A. Convolutional Layers

For convolutional layer, we first extract local patches from all training samples and form a local patch set $\mathbf{A} \in \mathbb{R}^{k_1 k_2 \times NP}$, where $[k_1, k_2]$ and $P$ denotes the size and number of local patch. $N$ is the number of training samples. Then, we compute the leading $L_1$ eigenvectors $\mathbf{V}_1 \in \mathbb{R}^{k_1 k_2 \times L_1}$ of $\mathbf{A}$ by PCA, and resize each column vector in $\mathbf{V}_1$ as the matrix of size $[k_1, k_2]$, which is considered as the filter kernel of the first stage. The convolutional response of the first stage can be obtained by convolving images with learned filter kernel. We further obtain the filter kernels of the $i$th stage by learning the leading $L_i$ eigenvectors of the local patch set of the $(i-1)$th stage's convolutional response. At last, we obtain the convolutional responses (i.e., feature maps) of all stages.

### B. Computation of Binary Feature Difference

Original PCANet produces a kind of discrete patterns' index instead of numerical features through *encoding*. It is meaningless to directly use a discrete index for second-order pooling [14]. In order to further improve the robustness of PCANet-II, we extend the *LBP difference* [14] to a more general BFD encoding scheme, so that non-numerical encoding output can be translated to the numerical one. BFD encodes binary features in any form, which are not limited to the particular form of local binary patterns. Moreover, we successfully apply BFD to the face analysis task, which is out of the scope of [14].

There are $L_i$ convolutional outputs $\{\mathbf{f}_l^i\}_{l=1}^{L_i}$ in the $i$th stage, where $L_i$ denotes the number of filter kernels in the $i$th-stage. We binarize these $L_i$ feature maps by the unit step function $\mathrm{S}(\cdot)$ and obtain the binary feature maps $\{\mathbf{B}_l^i\}_{l=1}^{L_i} = \{\mathrm{S}(\mathbf{f}_l^i)\}_{l=1}^{L_i}$. Therefore, the average pattern of the $l$th binarized feature map $\mathbf{B}_l^i(x, y)$ in the $i$th-stage can be defined as the mean as follow:

$$\mu_l^i = \frac{\sum\limits_{(x,y) \in \mathbf{f}_l^i} \mathbf{B}_l^i(x, y)}{N^i}, l = 1, 2, \cdots, L_i, \quad (1)$$

where $N^i$ denotes the dimensionality of $i$th-stage's feature map. The binary constraint is added to form the mean pattern in integer type:

$$\mu_l^i = \left\lfloor \frac{\sum\limits_{(x,y) \in f_l^i} \mathbf{B}_l^i(x, y)}{N^i} + 0.5 \right\rfloor, l = 1, 2, \cdots, L_i. \quad (2)$$

Having obtained the average pattern of each feature map, the BFD vector can be computed as $\left(\mathbf{B}_l^i(x, y) - \mu_l^i\right)$. We denote the binary feature difference with float mean and integral mean as BFD-F and BFD-I. The sign information is also used to encode BFD and form a discriminative descriptor as follow:

$$\mathbf{f}_{BFD}^i = \mathrm{sgn}\left(\|\mathbf{B}^i\| - \|\mu^i\|\right) \cdot \|\mathbf{B}^i - \mu^i\|, \quad (3)$$

where $\mathbf{B}^i = [\mathbf{B}_1^i, \mathbf{B}_2^i, \cdots, \mathbf{B}_{L_i}^i]$, $\mu = [\mu_1^i, \mu_2^i, \cdots, \mu_{L_i}^i]$, $\mathrm{sgn}(v) = 1$ when $v \geq 0$, and $\mathrm{sgn}(v) = 0$ otherwise.

We consider the BFD image $\mathbf{f}_{BFD}^i$ as a new kind of feature map. In each stage $i$, we cumulate this BFD image over the feature maps $\{\mathbf{f}_l^i\}_{l=1}^{L_i}$ to form the new feature map set $\mathbf{g}^i = [\mathbf{f}_1^i, \cdots \mathbf{f}_{L_i}^i, \mathbf{f}_{BFD}^i]$, so $\mathbf{g}^i$ has $L_i + 1$ column vectors.

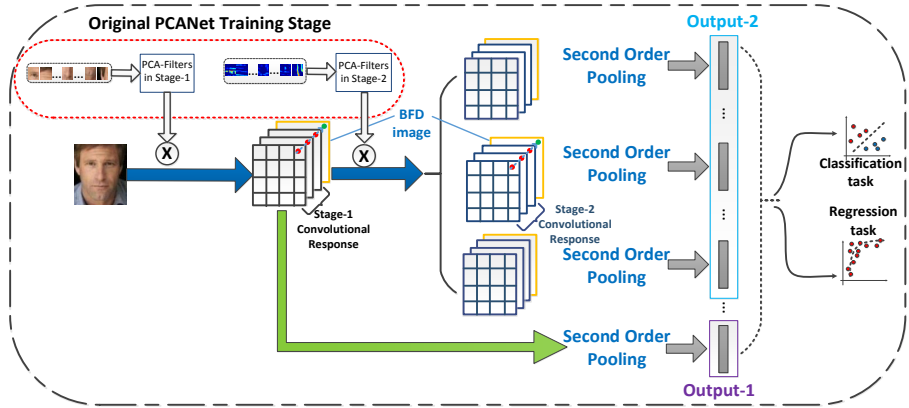

Fig. 1. The framework of our proposed PCANet-II method.

### C. Second Order Statistical Pooling

Compared with the histogram-based pooling, the second-order statistics can not only provide more discriminative information of feature maps by preserving the floating-point format, but also enable to integrate other informative cues (such as BFD) into the feature map set.

Since different local face patches have different configuration of facial components, we compute the patch-wise covariance matrices from a set of feature maps. More specifically, there are $(L_i + 1)$ output $\{\mathbf{g}_l^i\}_{l=1}^{L_i+1}$ in the $i$th stage, and each convolutional output $\mathbf{g}_l^i$ is divided as $M$ patches $\{\mathbf{g}_{l,m}^i\}_{m=1}^{M}$. The patch-based covariance matrix describes the $(L_i + 1) \times (L_i + 1)$ covariances between any pair of feature maps for the $m$th local patch. Its formulation is [15][16]:

$$CovM_m^i = c \times \sum_{l=1}^{L_i} \left(\mathbf{g}_{l,m}^i - \tilde{\mu}_m^i\right)\left(\mathbf{g}_{l,m}^i - \tilde{\mu}_m^i\right)^T, \quad (4)$$

where $c$ is the normalization constant and $\tilde{\mu}_m^i$ is the mean of $\{\mathbf{g}_{l,m}^i\}_{l=1}^{L_i+1}$. Therefore, the correlation of the local blocks of these "feature maps" is summarized by using the covariance matrix. Moreover, the covariance matrix has positive semi-definite and symmetric properties. So we just need around half the size of covariance matrix. Compared with the first order statistics, our second order pooling approach reduces the dimensionality of each block's output feature from $2^{L_i}$ to $(L_i + 1)(L_i + 2)/2$. At last, the covariance matrices of all stages are stacked as the final output vector.

## III. EXPERIMENTAL DATA AND SETUP

We choose three most representative datasets to investigate the performance of our proposed method for classification and regression tasks. For classification tasks, we use CAS-PEAL-R1 [17] and PaSC dataset [18] for constrained FR and unconstrained FR, respectively. For regression tasks, we use UNBC-McMaster pain dataset [19] for pain estimation (PE).

For CAS-PEAL-R1 dataset, we follow the evaluation protocols and use five subsets: *accessory*, *training*, *expression*, *gallery*, and *lighting*. Every image is aligned by provided eye coordinates and cropped into the size of $150 \times 130$. The PaSC dataset consists of 9,376 still images with large amounts of unconstrained variabilities, such as poor focus, lighting and motion blur. We align the all images according to the provided eye coordinates and crop into $128 \times 128$ pixels. There are two evaluation protocols: all image and near-frontal protocol. For the classification task, we use the Nearest Neighbor (NN) classifier with cosine metric. The UNBC-McMaster contains 48,398 frames and each frame corresponds to a PSPI score which can quantify pain intensity in 16 levels (0-15, 0 is the weakest and 15 is the strongest) [20]. We follow the pre-processing method as [21], and conducted leave-one-subject-out evaluation protocol. Different from the face classification task, the PE is a regression task. Therefore, we use the linear kernel Support Vector Regressors (SVR) (hyperparameter $C = 0.1$) [22] to obtain the estimated PSPI score. The average Mean Squared Error (MSE) is used to describe the performance of evaluated methods. The aligned and cropped examples of three datasets are shown in Fig 2.

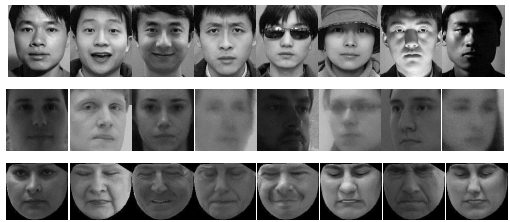

Fig. 2. The first row is CAS-PEAL-R1 examples. The second row is PaSC examples, the first four columns denote the near-frontal images and the last four columns denote non-frontal images. The third row is UNBC-McMaster examples. From left to right is: PSPI=0, 2, 4, 6, 8, 10, 12, 14.

According to experimental results, the performances of PCANet and PCANet-II are gradually improved with the increase of filter's number, while the running time and the output features' size are increased, too. Therefore, we set $[L_1, L_2] = [10, 10]$ for balancing the performance and computational cost. The filter size in the convolutional layer and local patch size in the pooling layer depend on the size of original face image. We always set the number of local patch to be $11 \times 11$ (i.e., $M = 121$) in these layers. The PCANet-II method with BFD-F and BFD-I are evaluated simultaneously. To make a fair comparison, we set the same parameters to existing PCANet method and follow the default parameters which are recommended by authors for other methods. The number of stages $i$ in our PCANet-II model is set to be 2, just like the original PCANet. At last, the Whitening PCA (WPCA) is used to further reduce the dimensionality of all methods under comparisons into $1,039$, $1,000$ and $1,000$ for CAS-PEAL-R1, PaSC and UNBC-McMaste datasets, respectively.

## IV. Experimental Results and Discussion

### A. Evaluation on CAS-PEAL-R1 Dataset

Table I provides the result comparison between our proposed and the state-of-the-art methods on CAS-PEAL-R1 dataset. As seen in Table I, our method achieved good results on the expression and accessory subsets and significantly outperformed all other methods on the lighting subset.

Compared with other state-of-the-art methods, the experimental results demonstrate that our method has excellent robustness and descriptive ability for various intraclass variabilities. Compared with original PCANet, our PCANet-II method not only preserves the magnitude information of feature maps by covariance matrix (corresponds to discriminative power), but also contains sign relationship of feature maps (corresponds to robustness) by BFD scheme. Because *Expression* and *Accessory* subsets are well controlled, our PCANet-II achieves slightly better accuracies than PCANet in these two saturated subsets. However, our PCANet-II significantly improves the accuracy of PCANet and other methods when faced with extreme lighting conditions.

TABLE I
THE ACCURACY (%) COMPARISON ON PEAL-R1 DATASET

| Methods | *Expression* | *Accessory* | *Lighting* |
|---|---|---|---|
| DFD [1] | 99.0 | 96.9 | 63.9 |
| IQBC [23] | **99.7** | **97.2** | 75.7 |
| E-LBP [24]* | 98.7 | 94.4 | 72.9 |
| PCANet [6] | 99.4 | 96.2 | 69.0 |
| **PCANet-II (BFD-F)** | 99.6 | 96.5 | **84.9** |
| **PCANet-II (BFD-I)** | 99.6 | 96.2 | 84.3 |

* We directly cite the results from the original papers.

### B. Evaluation on PaSC Dataset

Table II shows the verification rate at FAR=0.01 of our method and other state-of-the-art methods on PaSC dataset for all images and near-frontal images scenarios, respectively.

Our PCANet-II obtains good robustness by reference to the non-linear process in the PCANet model. Therefore, both of them achieve better performance when facing complicated intraclass variations. Compared with histogram-based pooling scheme of PCANet model, our PCANet-II retains more identity information (corresponds to discriminative ability) by computing the second order statistics of floating-based feature maps. Therefore, our method achieves the best performance on both all images scenario and near-frontal scenario. In other words, the second order statistical pooling provides more discriminative power than PCANet, and the BFD scheme brings our method good robustness.

TABLE II
VERIFICATION RATE (%) AT FAR=0.01 ON PASC DATASET

| method | all | frontal | method | all | frontal |
|---|---|---|---|---|---|
| LBP [9] | 17.6 | 29.6 | CBFD [25] | 19.4 | 36.0 |
| LRPCA [18]* | 10.0 | 19.0 | IQBC [23] | 21.2 | 38.8 |
| CohortLDA [18]* | 8.0 | 22.0 | PCANet [6] | 28.6 | 53.0 |
| BSIF [26]* | 14.3 | 24.9 | **PCANet-II (BFD-F)** | **30.4** | **54.0** |
| DFD [1] | 21.5 | 36.1 | **PCANet-II (BFD-I)** | 30.2 | 53.7 |

* We directly cite the results from the original papers.

### C. Evaluation on UNBC-McMaster Pain Dataset

We compare PCANet-II with the state-of-the-art PE methods as shown in Table III. Compared with those methods which are specifically designed for the PE task, the proposed general method achieved the second best MSE result. Even though MSE of our method is a bit worse than [16], our PCANet-II is learning-based instead of hand-crafted method, which makes it more data-adaptive and robust than the work [16] when faced with extreme intraclass variations. Though the PE is a regression task and is totally different from the recognition task, the essence of PE task is still the description and representation for facial image. From this view, the experimental results demonstrated the excellent facial description ability and robustness of our proposed model.

TABLE III
MSE COMPARISON ON UNBC-MCMASTER PAIN DATASET

| Methods | MSE | Methods | MSE |
|---|---|---|---|
| PTS [27] | 2.59 | 2Standmap [16] | **1.39** |
| DC [27] | 1.71 | VGGface+CNN+SVR [21] | 1.70 |
| LBP [27] | 1.81 | RCNN+Regression [21] | 1.54 |
| Gradient Histograms [28] | 4.76 | **PCANet-II (BFD-F)+SVR** | 1.47 |
| Hess+Grad [28] | 3.35 | **PCANet-II (BFD-I)+SVR** | 1.48 |

## V. Conclusion

In this paper, we propose a general "shallow" network for face analysis, named as PCANet-II. Compared with the histogram-based pooling method of PCANet, our model not only gains supplementary discriminative information by preserving both the magnitude and sign of convolutional responses, but also provides robustness by BFD scheme. Our proposed PCANet-II also solves the high dimensionality problem of histogram-based feature. Our general method achieves promising performances on both constrained and unconstrained FR and the PE tasks.

## Acknowledgment

This work was supported by the BUPT Excellent Ph.D. Students Foundation, the National Natural Science Foundation of China (Grants No. NSFC-61402046, 61572205), the Academy of Finland, Infotech Oulu, and Tekes Fidipro Program.

## REFERENCES

[1] Z. Lei, M. Pietikäinen, and S. Z. Li, "Learning discriminant face descriptor," *IEEE Transactions on Pattern Analysis and Machine Intelligence*, vol. 36, no. 2, pp. 289–302, 2014.

[2] J. Lu, V. Liong, G. Wang, and P. Moulin, "Joint feature learning for face recognition," *Information Forensics and Security, IEEE Transactions on*, vol. 10, no. 7, pp. 1371–1383, July 2015.

[3] F. Schroff, D. Kalenichenko, and J. Philbin, "Facenet: A unified embedding for face recognition and clustering," in *Proceedings of the IEEE Conference on Computer Vision and Pattern Recognition*, 2015, pp. 815–823.

[4] Y. Wen, K. Zhang, Z. Li, and Y. Qiao, "A discriminative feature learning approach for deep face recognition," in *European Conference on Computer Vision*. Springer, 2016, pp. 499–515.

[5] A. Krizhevsky, I. Sutskever, and G. E. Hinton, "Imagenet classification with deep convolutional neural networks," in *Advances in neural information processing systems*, 2012, pp. 1097–1105.

[6] T.-H. Chan, K. Jia, S. Gao, J. Lu, Z. Zeng, and Y. Ma, "Pcanet: A simple deep learning baseline for image classification?" *IEEE Transactions on Image Processing*, vol. 24, no. 12, pp. 5017–5032, 2015.

[7] Z. Lei, D. Yi, and S. Z. Li, "Learning stacked image descriptor for face recognition," *IEEE Transactions on Circuits and Systems for Video Technology*, vol. 26, no. 9, pp. 1685–1696, 2016.

[8] S. Yan, D. Xu, Q. Yang, L. Zhang, X. Tang, and H.-J. Zhang, "Discriminant analysis with tensor representation," in *Computer Vision and Pattern Recognition, 2005. CVPR 2005. IEEE Computer Society Conference on*, vol. 1. IEEE, 2005, pp. 526–532.

[9] T. Ahonen, A. Hadid, and M. Pietikäinen, "Face description with local binary patterns: Application to face recognition," *Pattern Analysis and Machine Intelligence, IEEE Transactions on*, vol. 28, no. 12, pp. 2037–2041, Dec 2006.

[10] C. Ionescu, O. Vantzos, and C. Sminchisescu, "Matrix backpropagation for deep networks with structured layers," in *Proceedings of the IEEE International Conference on Computer Vision*, 2015, pp. 2965–2973.

[11] T.-Y. Lin, A. RoyChowdhury, and S. Maji, "Bilinear cnn models for fine-grained visual recognition," in *Proceedings of the IEEE International Conference on Computer Vision*, 2015, pp. 1449–1457.

[12] P. Li, J. Xie, Q. Wang, and W. Zuo, "Is second-order information helpful for large-scale visual recognition?" *arXiv preprint arXiv:1703.08050*, 2017.

[13] K. Yu and M. Salzmann, "Second-order convolutional neural networks," *arXiv preprint arXiv:1703.06817*, 2017.

[14] X. Hong, G. Zhao, M. Pietikäinen, and X. Chen, "Combining lbp difference and feature correlation for texture description," *IEEE Transactions on Image Processing*, vol. 23, no. 6, pp. 2557–2568, 2014.

[15] O. Tuzel, F. Porikli, and P. Meer, "Pedestrian detection via classification on riemannian manifolds," *IEEE transactions on pattern analysis and machine intelligence*, vol. 30, no. 10, pp. 1713–1727, 2008.

[16] X. Hong, G. Zhao, S. Zafeiriou, M. Pantic, and M. Pietikäinen, "Capturing correlations of local features for image representation," *Neurocomputing*, vol. 184, pp. 99 – 106, 2016, roLoD: Robust Local Descriptors for Computer Vision 2014. [Online]. Available: http://www.sciencedirect.com/science/article/pii/S0925231215018949

[17] W. Gao, B. Cao, S. Shan, X. Chen, D. Zhou, X. Zhang, and D. Zhao, "The cas-peal large-scale chinese face database and baseline evaluations," *IEEE Transactions on Systems, Man, and Cybernetics-Part A: Systems and Humans*, vol. 38, no. 1, pp. 149–161, 2008.

[18] J. R. Beveridge, J. Phillips, D. S. Bolme, B. Draper, G. H. Givens, Y. M. Lui, M. N. Teli, H. Zhang, W. T. Scruggs, K. W. Bowyer *et al.*, "The challenge of face recognition from digital point-and-shoot cameras," in *Biometrics: Theory, Applications and Systems (BTAS), 2013 IEEE Sixth International Conference on*. IEEE, 2013, pp. 1–8.

[19] P. Lucey, J. F. Cohn, K. M. Prkachin, P. E. Solomon, and I. Matthews, "Painful data: The unbc-mcmaster shoulder pain expression archive database," in *Automatic Face & Gesture Recognition and Workshops (FG 2011), 2011 IEEE International Conference on*. IEEE, 2011, pp. 57–64.

[20] K. M. Prkachin, "The consistency of facial expressions of pain: a comparison across modalities," *Pain*, vol. 51, no. 3, pp. 297–306, 1992.

[21] J. Zhou, X. Hong, F. Su, and G. Zhao, "Recurrent convolutional neural network regression for continuous pain intensity estimation in video," in *Proceedings of the IEEE Conference on Computer Vision and Pattern Recognition Workshops*, 2016, pp. 84–92.

[22] A. J. Smola and B. Schölkopf, "A tutorial on support vector regression," *Statistics and computing*, vol. 14, no. 3, pp. 199–222, 2004.

[23] L. Tian, C. Fan, and Y. Ming, "Learning iterative quantization binary codes for face recognition," *Neurocomputing*, vol. 214, pp. 629–642, 2016.

[24] L. Liu, P. Fieguth, G. Zhao, M. Pietikäinen, and D. Hu, "Extended local binary patterns for face recognition," *Information Sciences*, vol. 358, pp. 56–72, 2016.

[25] J. Lu, V. E. Liong, X. Zhou, and J. Zhou, "Learning compact binary face descriptor for face recognition," *IEEE transactions on pattern analysis and machine intelligence*, vol. 37, no. 10, pp. 2041–2056, 2015.

[26] J. Ylioinas, A. Hadid, J. Kannala, and M. Pietikäinen, "An in-depth examination of local binary descriptors in unconstrained face recognition," in *Pattern Recognition (ICPR), 2014 22nd International Conference on*. IEEE, 2014, pp. 4471–4476.

[27] S. Kaltwang, O. Rudovic, and M. Pantic, "Continuous pain intensity estimation from facial expressions," in *International Symposium on Visual Computing*. Springer, 2012, pp. 368–377.

[28] C. Florea, L. Florea, and C. Vertan, "Learning pain from emotion: transferred hot data representation for pain intensity estimation," in *European Conference on Computer Vision*. Springer, 2014, pp. 778–790.